# Calibrated Trust in Dealing with LLM Hallucinations: A Qualitative Study

Adrian Ryser, Florian Allwein, Tim Schlippe
IU International University of Applied Sciences
Germany
adrian@ryser.eu, florian.allwein@iu.org, tim.schlippe@iu.org

*Abstract*— Hallucinations are outputs by Large Language Models (LLMs) that are factually incorrect yet appear plausible [1]. This paper investigates how such hallucinations influence users' trust in LLMs and users' interaction with LLMs. To explore this in everyday use, we conducted a qualitative study with 192 participants. Our findings show that hallucinations do not result in blanket mistrust but instead lead to context-sensitive *trust calibration*. Building on the *calibrated trust model* by Lee & See [2] and Afroogh et al.'s *trust-related factors* [3], we confirm *expectancy* [3], [4], *prior experience* [3], [4], [5], and *user expertise & domain knowledge* [3], [4] as user-related *(human) trust factors*, and identify *intuition* as an additional factor relevant for hallucination detection. Additionally, we found that trust dynamics are further influenced by *contextual factors*, particularly *perceived risk* [3] and *decision stakes* [6]. Consequently, we validate the *recursive trust calibration process* proposed by Blöbaum [7] and extend it by including *intuition* as a user-related *trust factor*. Based on these insights, we propose practical recommendations for responsible and reflective LLM use.

## I. Introduction

Large language models (LLMs) are reshaping information retrieval by enabling more conversational, context-sensitive search interactions [8]. Their dialogic interaction and high linguistic coherence make them an increasingly widespread tool in everyday use [9], [10]. At the same time, so-called hallucinations—i.e., LLM outputs that are factually incorrect but linguistically plausible—represent a key risk for users [11]. These are system-inherent properties of LLMs that are geared towards optimizing probabilities via token sequences [1], [12].

A global study by KPMG and the University of Melbourne [13] found that 66% of employees rely on LLM outputs without verifying accuracy. Similarly, the EY AI Sentiment Index 2025 [14] reports that fewer than one-third of users regularly verify AI-generated content. This suggests users often trust LLMs based on linguistic plausibility rather than verification (i.e., checking accuracy via external sources or domain knowledge)—especially when unaware of how outputs are generated [13]. This is problematic when users overestimate LLM capabilities [15].

Regardless of these issues, using LLMs requires a certain degree of a user's trust, not in the sense of blind acceptance, but as a conscious engagement with uncertainty. According to Lukyanenko et al. [16], human trust in AI is a process through which users adjust their interaction with a system based on how they perceive its capabilities (e.g., its functionality and limitations) and the context in which it is used. In practice, this means that a user with low AI literacy—i.e., insufficient ability to critically assess how LLMs function and where their limitations lie [17]—must rely more heavily on trust to mitigate uncertainty during interaction. Conversely, a user with high AI literacy can identify error sources (e.g., hallucinations) early and engage more reflectively with LLMs, reducing the need to rely on trust as a bridge over uncertainty [18].

Previous studies have examined how response structure, the presence of source references, or text which sounds convincing influence user's trust in LLMs—typically in controlled experiments with predefined prompts [19], [20], [21]. However, users' personal experiences, individual strategies, and the everyday use of LLMs have not yet been examined. Moreover, there is a lack of research which tackles how LLM hallucinations influence users' trust [22].

To address these gaps, we conducted a qualitative study with 192 participants to answer the following research question: **How do experiences of LLM hallucinations influence users' trust in LLMs and users' interaction with LLMs?**

Our contributions are as follows:

- Empirical insight into user trust: In our qualitative study, we analyze how users perceive and respond to hallucinated LLM outputs, showing how trust is contextually calibrated and managed under uncertainty.
- Extension of the *calibrated trust model* by Lee & See [2], Afroogh et al.'s *trust-related factors* [3] and Blöbaum's *recursive trust calibration process* [7] to the context of hallucination-prone LLM usage based on our findings.
- Recommendations for LLM users: (1) calibrate trust, (2) verify contextually, (3) integrate intuition into trust assessment, (4) build AI literacy, and (5) treat LLMs as assistants.
- Region-specific empirical contribution: To the best of our knowledge, this is the first qualitative study on LLM hallucinations and trust in the German-speaking DACH region.
- Open data: Anonymized responses are available at: https://doi.org/10.5281/zenodo.15618622

## II. RELATED WORK

### A. Hallucinations in LLMs

Hallucinations are defined as content generated by LLMs that is linguistically convincing but factually incorrect [1], [12]. Studies report a high rate of hallucinations in LLM outputs, particularly in citations and medical content [23], [24]. Root causes of such hallucinations include flawed or biased training data, a misalignment between training objectives and *user expectations*, as well as limitations in decoding and inference strategies [22]. Given the highly convincing and human-like responses generated by LLMs, detecting hallucinations is particularly difficult for users in practice [22].

Despite efforts to mitigate hallucinations, e.g., specific prompting techniques [25] and model-level improvements [22], recent theoretical work shows that hallucinations cannot be fully eliminated, as no model can produce factually correct outputs for all possible inputs [26], [27]. This highlights the need to better understand user behavior and verification strategies in response to hallucinations [22]. To date, research has focused largely on model-level mitigation, with less attention to end-user verification and response patterns.

### B. User reactions and verification behavior

Despite the risk of hallucinations, users rarely verify LLM outputs systematically. According to the EY AI Sentiment Index [14], fewer than one in three users verify AI-generated content, whereas in Germany, only 27% do. A global study by KPMG and the University of Melbourne [13] found that 66% of employees trust LLM outputs without checking them—and over half reported work-related mistakes due to over-reliance on LLM outputs.

This indicates that even when verification is possible, many users trust LLM outputs by default without questioning their accuracy. Consequently, verification behavior is closely intertwined with users' underlying level of trust in the LLM. Studies also show that trust and verification behavior differ by application context: For instance, Mendel et al. [21] found that users perceive LLMs as less reliable than traditional search engines for health-related questions.

Verification decisions appear to depend on perceived relevance and potential consequences. However, most existing studies focus on predefined tasks in controlled settings, offering limited insight into users' subjective experiences with LLMs in everyday use [28], [29]. Further research is needed to examine how hallucinated outputs affect users' perception of uncertainty, verification behavior, and trust in everyday LLM use.

### C. Trust under uncertainty

As shown in Section B, users often rely on LLM outputs in practice, even though they rarely verify them systematically [13], [14]. Trust often compensates for the absence of verification, particularly under conditions of uncertainty, time pressure, or cognitive load [29], [30]. Here, trust serves as a cognitive bridge between limited knowledge and action-oriented decisions [16], [30].

Lukyanenko et al. [16] describe human trust in AI as a mental and physiological process through which users adjust their interaction with a system based on how they perceive its capabilities and the context in which it is used. To better understand how such trust emerges, it is helpful to revisit traditional definitions of trust: Luhmann [18] defines trust as a central mechanism for reducing uncertainty—both in a social and technical context. According to him, trust is a strategy for decision-making under incomplete information. Rousseau et al. [31] stress trust's emotional component by stating it is the "willingness to be vulnerable to an actor who is expected to behave benevolently".

While these definitions provide a conceptual foundation, empirical studies on trust in LLMs in controlled experimental settings show that user perception can often be shaped more by surface-level features than by factual accuracy [10], [15], [19], [20]. Users tend to trust LLM output that contains references, even when those sources are fabricated or irrelevant. Trust typically declines when such references are verified and found to be inaccurate [19]. Well-structured and fluent outputs are also perceived as more trustworthy, regardless of their factual validity [20]. Linguistic coherence and a confident tone can create an illusion of competence, leading users to overestimate the reliability of LLM outputs [15].

Trust has been identified as a key factor in user engagement—even in the absence of systematic verification [10], [15]. This suggests that trust is shaped not only by system performance but also by presentation, *user expectations*, and context. Under uncertainty, it functions as a factor that enables decision-making without complete information.

### D. Trust-related factors

Human-AI interactions are characterized by a structural asymmetry of information: Users cannot fully comprehend the system logic, but must place a certain degree of trust in it to be able to use the system [16]. Lee and See's *model of calibrated trust* [2] addresses this by proposing that trust should align with the system's actual capabilities and limitations, rather than be granted unconditionally. The model distinguishes three trust states: *Undertrust*, originally termed *distrust* by Lee and See [2], refers to insufficient reliance on the system despite its actual capabilities. In the LLM context, the term *undertrust* is preferred to highlight missed opportunities due to overly cautious behavior [3], [5]. *Overtrust* describes excessive reliance beyond the system's reliability, often leading to uncritical acceptance of errors such as hallucinations. *Calibrated trust*, by contrast, reflects an appropriate alignment between user reliance and actual system performance. Both *overtrust* and *undertrust* carry risks: While *overtrust* may lead to the uncritical acceptance of incorrect outputs like hallucinations, *undertrust* can prevent effective use of LLMs. Therefore, users should aim for *calibrated trust*. A visualization of the factors influencing *trust calibration* in LLM interaction is given in Figure 1.

*Calibrated trust* is influenced by several factors [4], [5], [6], [32], [33]. *Trust factors* which influence *trust calibration* are for example *user expertise & domain knowledge* (familiarity with the subject matter), *prior experience* (experience with AI tools like LLMs), *expectancy* (anticipated system performance), *perceived risk* (subjective anticipation of harm), and *decision stakes* (objective relevance or consequences of a decision). Some *trust factors* remain debated or context-dependent: *anthropomorphism* may foster *overtrust* or *undertrust* depending on *user expectations*, while *demographic* factors such as age and education appear neutral or inconsistent across studies [3], [4], [6]. Afroogh et al. [3] map *trust factors* into four categories: *technical*, *human*, *contextual*, and *axiological*.

*Intuition* (fast, experience-based evaluations [34]) has also been found as a factor to influence AI users' decisions by Chen et al. [36]. They investigated user interactions in AI decision-support systems that provide explicit explanations and found that users appear to rely on their *intuition* to decide when to trust or override AI predictions. However, they investigate neither LLMs nor the impact of hallucinations. Therefore, it remains unclear how *intuition* influences *trust calibration* when users encounter hallucinated LLM outputs.

*E. Calibrated trust as a dynamic process*

Trust in LLMs should not be seen as a static state, but as a dynamic process that evolves in response to ongoing *user experiences*. As Blöbaum [7] emphasizes, trust forms through a recursive process in which past interactions shape future trust judgments—especially in uncertain or unfamiliar situations. This understanding of trust as a recursive process [7] complements the concept of *calibrated trust* by highlighting its temporal nature.

*F. Research gaps*

While the concept of *calibrated trust* is theoretically well established, empirical insights into how users adjust their trust in response to LLM hallucinations remain limited. Most existing studies on trust in LLMs, including those addressing hallucinations [19], [20], [21], [36], are conducted with predefined tasks (e.g., answering fixed prompts) or in controlled experimental settings. However, since trust is dynamic and context-dependent, its development in everyday LLM use under the risk of hallucinations is still poorly understood. Furthermore, little is known about how *intuition* affects *trust calibration*. In addition, there is a lack of research on LLM trust in German-speaking contexts. A qualitative approach is needed to explore which trust factors shape calibration processes during everyday LLM use in the presence of hallucinated content.

## III. METHODOLOGY

*A. Methods for data collection and analysis*

Following Blöbaum [7], we see trust as a dynamic process, not a measurable quantity. Consequently, we analyzed trust using qualitative methods [37]. Our aim was to find out how experiences of LLM hallucinations influence users' trust in LLMs by researching subjective perspectives. We used a qualitative online survey as the method of data collection in order to combine the openness of qualitative data collection methods with wide reach and flexibility in terms of time [38]. This allowed participants to answer the survey at their own time rather than putting them in an artificial setting like a lab. We analyzed data using an iterative, theory-driven approach that combined elements of the hermeneutic circle and Mayring's qualitative content analysis [39]. Coding was conducted by one of the authors and double-checked by another author. Given the high number of responses, theoretical saturation was reached for our categories. Responses were coded using categories like "context of LLM use", "perception and handling of hallucinations" or "measures to deal with hallucinations". We developed these further using a systematic coding process, supported by the Atlas.ti software [40].

We ensured scientific rigor in our data collection and analysis by following Mayring's quality criteria [39]. This included a systematic and rule-guided procedure of data analysis as well as closeness to participants' world in the survey (which they could answer in their own time and in their usual environment). We carefully documented the process of data analysis and continuously refined the categories derived.

*B. Survey*

We conducted a qualitative online survey between December 6, 2024 and January 8, 2025, asking participants about their experiences using ChatGPT as an example of LLMs. We followed IU International University of Applied Sciences' guidelines for ethical research. Participants were asked for consent to have their data used. We collected responses without personally identifiable information, unless participants volunteered to share it.

Participants were initially asked about how often they use LLMs. Participants who never used it (n=5) were only asked for the reasons, but not any other questions. The remaining participants were asked a combination of 4 multiple choice and 11 open ended questions (none of which were mandatory). After a pretest, the final questionnaire was provided to participants. We asked questions like whether participants knew hallucinations existed, whether they had observed any hallucinations before, whether this affected their trust in LLMs and whether they used LLMs as a supporting tool or a primary source of information. Even with Yes/No questions like these, participants were given an opportunity to comment on their answer or give more details about their experiences.

192 people participated (66% female, 32% male, 1% other, 1% undefined) in the online survey and provided complete answers. The majority of participants (40%) were between 26 and 35 years old, with the age spectrum ranging from 18 to over 55. 65% lived in Germany, 29% in Switzerland, 6% in Austria or other countries. 36% used LLMs daily, 43% weekly, 18% monthly or less frequently; 3% stated that they had never used LLMs specifically. The majority (66%) used a free version, while 34% used a paid subscription. 26% were students at the time of the survey, the majority were employed.

Participants were recruited through our university's communication channels, LinkedIn and e-mails. The high number of participants (192) compared to other qualitative studies, as well as their variety, allow us to assume that our results are representative for other persons from similar cultural backgrounds.

## IV. FINDINGS

To support understanding, we include selected examples from participant responses throughout the text. These responses were translated into English by one of the authors using a semi-automated process. They were then anonymized and labeled with codes such as A1, A2, etc., where "A" stands for "Answer." The original quotes are available in our repository (see Sec. I).

*A. Context of use*

Five participants (3%) reported that they do not use LLMs and consequently were not asked subsequent questions. The reasons they gave were mistrust, lack of awareness, or value-driven reasons. Importantly, this mistrust was expressed before participants were shown the hallucination example and was therefore not influenced by it. These reasons ranged from fundamental skepticism to a lack of trust in the quality of the answers ("I don't trust the quality", A133). Value-based motives were also mentioned—such as the desire for independent thinking or not using technical aids: "You are allowed to use your brain more yourself" (A129).

All results in the following sections refer to the remaining participants (n = 187). They used LLMs in various ways, especially for brainstorming, improving writing, and exploring new topics. As an example, A32 described using LLMs "for all areas of my life." A total of 19 participants stated that LLMs are replacing traditional search engines for them (cf. A16, A45, A136, A185).

Some participants also described specific restrictions on use, particularly in an academic context: "I don't use it for my studies either because I want to do my work myself" (A4). Overall, there is widespread everyday use, although LLMs are deliberately avoided or only used to a limited extent in certain contexts like the aforementioned.

*B. Perception and handling of LLM hallucinations*

Most participants (82%) were familiar with the concept of LLM hallucinations. About two thirds (68%) reported personal experience, while 18% denied such experience, and 14% were unsure whether they had already been affected by LLM hallucinations. Among those with concrete experiences in LLM hallucinations (68%), more than half (58%) described poor content quality as the key feature for identifying them. Additionally, 5% reported relying on *intuition* (fast, experience-based evaluations) [34] and *common sense* (the application of everyday knowledge to detect implausible content) [35] to detect hallucinations. Typical examples of this perception included: "They contradicted common sense" (A88) and "Answers were inconsistent or simply wrong" (A178).

Some participants (14%) also expressed uncertainty about whether they had experienced hallucinations. This uncertainty was attributed to infrequent use, low awareness of LLM hallucinations, or specific forms of use that were perceived as less prone to hallucinations. For example, participant A54 noted: "I haven't noticed them, but I can't rule out being affected". Others described their LLM use as relatively safe, such as A53: "I have texts shortened or sentences rearranged that I have written myself. If I don't like the result, I don't use it or only use parts of it".

*C. Effects of LLM hallucinations on trust*

Participants reported varying effects of hallucination experiences on their trust in LLMs, ranging from no change to complete loss of trust. While 49% stated that their trust remained unchanged, 20% reported a minor decrease in trust, and 31% described a significant loss of trust.

*Unrestricted trust* in LLMs was rare and was explicitly expressed in only six cases (3%). For example, participant A164 stated: "However, as I usually get very good answers, the level of trust is still very high. As a percentage, I would put it at 90-95%". More commonly, participants described a form of differentiated trust that combined general confidence with critical awareness. One participant put it as follows: "Great trust with a healthy degree of skepticism" (A17). Others expressed initial caution ("I was very skeptical from the beginning," A23) or a clear sense of mistrust: "Little trust, especially with important topics" (A64).

One of the core categories we developed inductively was *trust according to relevance*. This refers to participants' tendency to adjust their level of trust depending on the perceived relevance or consequences of a given use case. Some participants (11%) reported consciously adapting their level of trust depending on situational factors such as during their studies, at work, or when making significant decisions. Participant A17 explained: "If it's ok to be wrong, then I trust the information 100%, but if it has to be correct, then it's more like 70-80%". A32 similarly emphasized the situational nature of trust: "Depends on the situation, [...] for important and complex questions, especially in areas where I have no knowledge myself, I would never trust ChatGPT. [...] For everyday questions or purchasing decisions, however, I rely very much on the accuracy of the answer."

Among participants who reported a significant loss of trust (31%), different causes emerged: Some, like A56 and A99, were not previously aware of hallucinations and likely lost trust during the survey, after seeing the example hallucination. A56 noted: "Looking at the example, now every answer AI has ever given me makes me feel insecure." A99 added: "Damn, now I have even less trust. Especially because I had a lot of texts summarized. Let's see if I'll stop using ChatGPT altogether." Others, such as A171 and A180, already knew about hallucinations and described a gradual decline in trust over time. A171 stated: "I've started to critically question things much more" and A180 recalled: "Before, I actually thought almost naively that there was a relatively high probability that it would answer all my questions correctly."

Additionally, some participants (9%) expressed emotional reactions in response to their experiences with LLM hallucinations. These ranged from frustration ("I was rather annoyed", A178) to concern ("dangerous source of false information", A160) to irony ("People should continue to think for themselves [...] just like in real life ;-)", A107). Such statements highlight how trust in LLMs is not only a cognitive judgment about information reliability but also emotionally charged, especially when participants feel personally affected by hallucinations or unreliable outputs.

Participants' usage behavior changed in response to their experiences with LLM hallucinations. Two participants reported reducing their use of LLMs as supporting tools. As A30 noted: "Since I've known about the hallucinations, [I use ChatGPT] only as a supporting tool." Others described a shift back to more established or traditional sources of information: "[I use ChatGPT] less as a source of information, I switched back to normal 'googling'" (A9). Some continued using ChatGPT but with increased caution. A28 noted: "It made me a bit suspicious, but I still don't use ChatGPT any less." A few participants put the risk of hallucinations into perspective by drawing comparisons to human behavior, as in A37's remark: "[...] and it can make mistakes like a human."

*D. Output verification*

Participants reported different strategies for verifying LLM outputs. More than one third (38%) stated that they always verify the information they receive, while about half (51%) reported checking outputs selectively, depending on the situation. In contrast, a smaller group (11%) said they do not verify the content at all. Several participants reported consistently verifying LLM outputs, particularly when using them in academic or professional contexts. A3 emphasized a general lack of reliability: "ChatGPT always makes things up and you can't be sure whether they are actually true." Others mentioned the need for additional sources in order to form balanced judgments. As A152 explained: "[Y]ou need information from several sources in order to check it. Otherwise, it can be too one-sided."

Analogous to the category *trust according to relevance* (see Sec. IV C), we identified an inductively derived category *output verification according to relevance*. One quarter (25%) of participants described a context-sensitive approach to verification, based on expected consequences of potential hallucinations. These decisions—to verify or not—were influenced by a range of *trust-related factors*, including *perceived risk*, such as the fear of relying on incorrect information [3], and *decision stakes*, referring to the objective relevance or consequences of a particular output [6]. Both factors are well-established as *contextual factors* in trust research (see Fig. 1). In addition, *prior experience* [3], [4], [5] and *domain knowledge* [3], [4] played a role in shaping users' willingness to verify information. Some participants also mentioned relying on *intuition* when judging the plausibility of an answer. A25 explained: "I only verify them if the answers seem wrong or unclear to me", while A8 described this *intuitive* approach more explicitly: "If the answers seem strange to me, I follow my gut instinct". While *intuitive* judgments are not explicitly included in existing trust models, they were mentioned by several participants as a basis for deciding whether or not to verify information.

In academic or professional contexts, *perceived risk* often resulted in verification: "[...] if it really has to be correct, for example if I need it for a university essay, then I like to verify the information again myself" (A17). In contrast, verification was often deliberately omitted in non-critical contexts, such as when participants searched for simple facts, general knowledge, or personally irrelevant information (A21, A49, A72, A103). A32 described a nuanced approach to output verification, combining several factors in their decision-making. These included the potential impact of an incorrect answer, their own ability to assess the information, and the perceived logic of the response. As A32 explained: "The greater the impact of an incorrect answer, the less I can check the correctness of the answer using my own knowledge and the more illogical the answers seem to me, the more likely I am to verify the answers."

Google itself was occasionally mentioned as a supplementary tool for checking information (A28, A40, A85). However, some participants also expressed skepticism toward this strategy: "Google is not always right either" (A72).

Only 3% of participants explicitly justified why they did not verify LLM outputs. In most cases, the lack of verification appeared to stem from missing *user expertise & domain knowledge*. For example, A113 explained: "I thought ChatGPT was reputable", and A137 added: "I trust that the information was taken from valid Google sources [...]"

### E. Measures suggested by users

Participants made a range of suggestions for reducing LLM hallucinations, which were grouped into two inductively derived categories:

#### 1) System-related measures

One in five participants (20%) suggested a consistent citation of sources by LLMs to improve the transparency of responses. Several also proposed implementing a confidence score to indicate reliability. A162 called for: "A kind of score could be displayed in the result with which confidence ChatGPT has produced this result." and A25 stated: "Some kind of probability score would be helpful." Additionally, 16% recommended including clear warnings about the uncertainty of generative models. For example, A65 proposed: "A prominent indication that hallucinations may occur." Some also called for more explainable artificial intelligence (XAI), such as revealing the underlying data or making the reasoning behind an answer traceable (A59, A149). A few participants also suggested improving the user interface, for example by providing clearer disclaimers about the potential for LLM hallucinations (A43, A156).

#### 2) User-related approaches

Some participants (18%) proposed training programs, guidelines, or usage standards for the users, particularly in professional settings. These included suggestions such as mandatory verification protocols in companies (A136). A recurring theme was the importance of AI literacy (A173). Some participants also emphasized the role of individual reflection: For example, they warned against user convenience ("because we are simply too lazy", A45) or appealed to common sense ("Switch on your brain [...] get out of your comfort zone and think more for yourself", A159). Other participants explicitly mentioned using *common sense* to question LLM outputs (A115, A163).

## V. ANALYSIS AND DISCUSSION

### A. Answering the research question

We investigated how hallucinations influence users' trust in LLMs and users' interaction with LLMs. Hallucination experiences did not lead to a general loss of trust. As discussed in Section IV.C, while some participants of our study reported a decrease in trust, most described changes in how they interact with LLMs. Trust was not lost, but recalibrated based on *prior experience* and the perceived relevance of the task.

Participants reported a widespread use of LLMs in everyday situations to support thinking processes and simplify tasks, particularly for brainstorming, structuring texts, and language refinement, as outlined in Section IV.A. In some cases, traditional search engines were partially replaced by LLMs. However, in contexts such as academic work or professional tasks, where accuracy is critical, users often deliberately limited their use of LLMs.

Participants employed various forms of LLM output verification—especially when they considered the consequences of possible hallucinations to be serious. As outlined in Section IV.D, this verification was context-dependent: more rigorous in academic or work-related contexts, and minimal or entirely absent in everyday or low-risk use cases. Additionally, several participants reported relying on *intuition*—sometimes expressed as *common sense* judgments—when assessing the plausibility of LLM outputs, particularly when verification through external sources (e.g., Google) was not feasible. Since *intuition* emerged as a distinct and recurring element in participants' descriptions, we will consider it further as an additional *trust-related factor* in the next subsection.

This differentiated use of LLMs, shaped by user awareness, output verification, and the use of *intuition*, points toward a dynamic and context-sensitive trust mechanism. Rather than trusting LLMs unconditionally or rejecting them entirely, participants reported adapting their trust based on *prior experience*, *perceived risk*, and *decision stakes*. These findings are consistent with the concept of *calibrated trust*, which we will examine in more detail in the following subsection.

## B. Influencing factors of calibrated trust

Lee and See's model of *calibrated trust* [2] suggests that trust should match system reliability across contexts. Our findings support this view and show how users *calibrate trust* based on LLM hallucination experiences. Consistent with prior literature (see Section II), our survey's results confirm known *trust factors* like *expectancy*, *user expertise & domain knowledge*, *prior experience*, *perceived risk* and *decision stakes*.

In Figure 1, we extended Lee & See's *calibrated trust model* [2] with *trust-related factors* in LLM interactions, categorized into *technical*, *axiological*, *contextual* and *human factors* as suggested by Afroogh et al. [3]. Arrows indicate possible influence directions between *overtrust*, *undertrust* or *calibrated trust* based on our interpretation. The *trust factors'* references are indicated in IEEE citation style.

As demonstrated in the figure, our findings suggest *intuition* as a factor that may support the early detection of implausible or hallucinated content through quick judgments about whether an LLM output seems plausible. Bold text in Figure 1 indicates factors supported by our survey's findings; *intuition* (red) emerged as a new factor in the context of LLM hallucinations.

While *intuition* draws on *prior experience*, it functions differently: It refers to fast, unconscious judgments based on pattern recognition and familiarity, not conscious reasoning based on *prior knowledge*. This distinction reflects Kahneman's *dual-process theory* [34], where System 1 supports rapid, experience-based reasoning (i.e., *intuition*), and System 2 underpins slow, reflective analysis. Kahneman argues that these two systems both serve valuable purposes and complement each other. As outlined in Section II.D, *intuition* in LLM use takes various forms—such as quick judgments based on linguistic coherence, internal consistency, or output logic [36]. Participants frequently described relying on such judgments—especially when *domain knowledge* was unavailable or verification of the LLM output was not feasible. We argue that *intuition* complements rather than duplicates *prior experience*, acting as a real-time plausibility filter—particularly useful in time-sensitive or low-stakes situations where verification of LLM output is unlikely.

At the same time, as Kahneman [34] points out, fast, experience-based judgments like *intuition* are also prone to systematic biases and an "illusion of validity", which can foster unwarranted confidence in unreliable outputs. This highlights the ambivalent role of *intuition*: while it can support rapid plausibility checks, it may also mislead users in contexts that require accuracy or expertise.

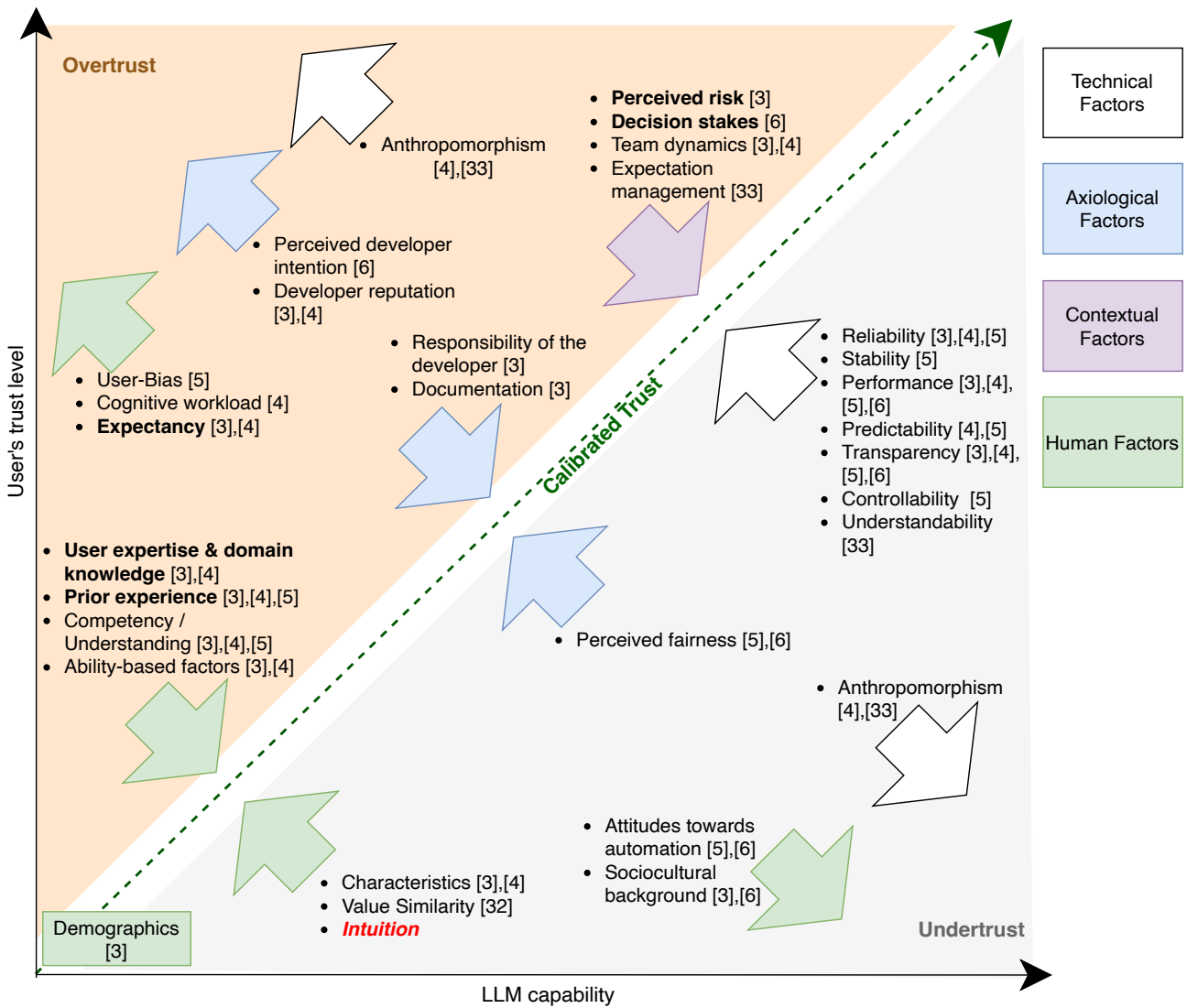

**Fig. 1** Extended *calibrated trust model*: Factors influencing *trust calibration* in LLM interactions.

The influencing factors of *calibrated trust* displayed in Figure 1 are valid under the condition that users are aware of LLM hallucinations (*cognitive factors*). However, our results show that there are conditions where users were unsure whether they had encountered LLM hallucinations at all. This points to a limitation of *calibrated trust*: When *prior knowledge* is low and *intuitive* cues are weak, hallucinations may go undetected, and misplaced trust may persist. A small group of users responded to this uncertainty with deliberate non-use, either generally or within specific domains. Motivational factors included skepticism, personal values, and a desire for cognitive independence. In these cases, *trust calibration* did not lead to adaptation but rather to withdrawal from LLM use, highlighting the limits of our extended *calibrated trust model*.

Beyond these cases of withdrawal, our findings also revealed internally inconsistent attitudes toward trust. Some participants expressed skepticism in most use cases but trust for one specific use case. For example, they reported always verifying LLM outputs while at the same time, they admitted to rely uncritically on LLM-generated summaries. Such contradictions suggest that *trust calibration* does not follow a linear or uniform process, but can vary depending on situational factors such as time pressure or task complexity. This underscores the need to conceptualize *calibrated trust* as a dynamic and sometimes ambivalent process.

In addition to *cognitive factors* such as the awareness of LLM hallucinations, *emotional factors* also shape users' trust. As discussed in Section IV.C, emotional responses played a role in how participants interpreted their experiences with hallucinated LLM outputs. This highlights that trust is not solely based on rational evaluation but also influenced by emotional experience. Afroogh et al. [3] include such emotional aspects with regards to human-machine interaction in their review article, but do not focus on LLM hallucinations.

### C. Recursive trust calibration in hallucination-prone LLMs

As shown in Figure 1, *trust factors* are not static—they evolve through repeated user interaction with LLMs.

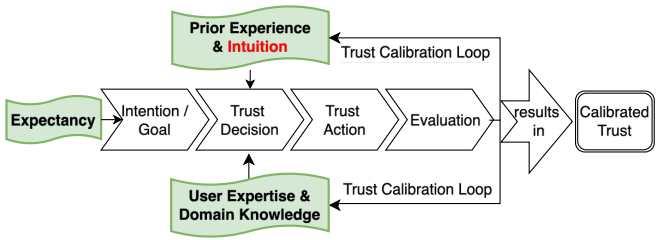

**Fig. 2** *Trust calibration* process in LLM interactions under perceived risk and decision stakes in hallucination-prone-LLM interactions [7].

Based on Blöbaum's [7] *recursive trust calibration process*, we adapted and expanded this process to the case of hallucination-prone LLMs. Figure 2 presents this iterative process, illustrating how users gradually develop *calibrated trust* through practical experience with given LLM outputs. The core process consists of four sequential phases: (1) *Intention/Goal*—The user initiates interaction with a purpose or expectation. (2) *Trust decision*—Based on influencing factors, the user decides whether to trust the system. (3) *Trust action*—The user acts on this trust by engaging with the system. (4) *Evaluation*—The user reflects on the outcome, leading to a potential adjustment of trust. Three key *user-related factors* influence this process: (1) *Expectancy*—Initial assumptions about the system's capabilities. (2) *Prior experience*—Previous encounters with similar systems. (3) *User expertise & Domain knowledge*—Background knowledge that supports judgment.

Our study extends this *recursive trust calibration process* by identifying *intuition*—highlighted through our qualitative findings—as an additional *user-related trust factor* in LLM interactions. In addition to user-related factors, trust dynamics are shaped by *contextual factors*, particularly *perceived risk* [3] and *decision stakes* [6], which determine the relevance and potential consequences of hallucinated LLM outputs. While not visualized in Figure 2 for clarity, these factors play a key role in the *trust calibration process*.

Over time, each *trust calibration loop* leads to increased *AI literacy*—understood here as the ability to engage with LLMs critically, appropriately, and with an awareness of their limitations, depending on the use context [17]. Consequently, *calibrated trust* can be understood as a user competence shaped by various *cognitive* and *behavioral factors*—among them, *intuition*—which together support effective interaction with LLMs, even under conditions of uncertainty.

### D. Recommendations for users

Since LLMs cannot reliably indicate their own uncertainty [15], [41], the responsibility for interpretation and verification remains largely with the user. The concept of *calibrated trust* provides a foundation for this user responsibility: It enables users to avoid both uncritical acceptance and blanket rejection of LLM-generated content. Drawing on our findings, we propose five principles for informed and reflective interaction with hallucination-prone LLMs:

1) **Calibrate trust**: Users should actively *calibrate their trust* in LLM outputs by considering the task's relevance and their own level of *domain knowledge*.
2) **Verify contextually**: By tailoring verification efforts to the *perceived risk* and relevance of the task, users can manage uncertainty efficiently.
3) **Integrate intuition into the trust assessment**: Based on *prior experience* with LLMs, users should rely on their *intuition* for linguistic coherence, internal consistency, or implausibility to detect hallucinated LLM outputs—especially when output verification is not feasible.
4) **Build AI literacy**: Developing a better understanding of how LLMs function and where their limitations lie is essential for responsible and context-sensitive use.
5) **Treat LLMs as assistants**: A clear classification of LLMs as a supporting tool promotes an appropriate approach to the limits of LLMs.

### VI. CONCLUSION AND FUTURE WORK

#### A. Contributions to research

Our study with 192 participants contributes to understanding how hallucinations in LLMs affect users' trust, and which strategies they use to adjust, stabilize, or recalibrate it. To the best of our knowledge, it is the first qualitative-empirical study in the German-speaking world to address this issue with users in daily use scenarios. LLMs are predominantly used as supportive tools by the participants—particularly for idea generation, structuring, or language refinement—and are rarely used as the sole source of information. While outputs that are professionally or personally significant are often verified, responses to everyday questions or minor tasks are typically accepted without further verification.

Trust in LLMs is not a fixed state, but a dynamic, experience-based process. Participants learn to assess the plausibility of outputs, adapt their *expectations*, and apply context-sensitive verification strategies. *Intuition* supports this process, particularly in situations where LLM output verification is not feasible or when *perceived risk* of the current task is low.

*B. Limitations*

Our qualitative survey enabled broad, flexible participation and rich, experience-based insights. However, the survey also has some limitations. The study offers only a single time-point snapshot. Longitudinal work could track evolving trust over time. While pre-testing the questions in the survey led to successful answers, suggesting overall comprehension, misunderstandings cannot be ruled out. As a qualitative study, the emphasis lies on interpretation rather than statistical generalization, even though we report descriptive numbers from closed questions and coded responses to illustrate tendencies.

Demographic data such as age, gender, occupation, subscription status, and frequency of LLM use were collected but not systematically analyzed for subgroup differences, as this was not relevant for our research questions. In addition, some participants reported a loss of trust only after being confronted with the example LLM hallucination in the survey itself (e.g., A56, A99). This indicates that parts of the findings may reflect reactive awareness induced by the study rather than solely prior experience. Future research with larger and more diverse samples could validate and extend these insights.

*C. Future research*

This paper opens up a number of points for future research:

- It would be interesting to triangulate our survey results with objective behavioral sources such as chat logs or system metrics, to deepen understanding of LLM use.
- A central question is how trust evolves over time. Since our data reflects a single point of observation, longitudinal studies are needed to explore how repeated use, software updates, and educational interventions affect trust dynamics.
- A broader and more diverse set of participants may reveal trust differences across age, education, cultural background and technical familiarity, e.g., AI literacy.
- Since some participants contradicted their statements (e.g., accepted LLM-generated summaries uncritically despite stating to verify LLM output completely), further analyzing users' trust deserves particular attention.
- Emotional responses such as frustration or irony suggest that trust in LLMs is shaped not only cognitively, but also emotionally, which requires further investigation.

Finally, our study shows that there is only limited awareness of hallucinations among users of LLMs. Given that hallucinations are an inherent property of LLMs with potentially significant negative consequences, we call on researchers and users of AI and LLMs to be mindful of them and adopt their use accordingly.